\documentclass[runningheads]{llncs}
\usepackage{graphicx}
\usepackage{hyperref}
\usepackage{academicons}
\usepackage{xcolor}
\usepackage{enumitem}
\usepackage{subcaption}
\usepackage[nobiblatex]{xurl}
\usepackage{seqsplit}
\usepackage[TU]{fontenc}
\usepackage{orcidlink}

\PassOptionsToPackage{hyphens}{url}\usepackage{hyperref}

\usepackage{academicons}
\definecolor{orcidlogocol}{HTML}{A6CE39}

\newcommand{\orcid}[1]{\href{https://orcid.org/#1}{\textcolor[HTML]{A6CE39}{\aiOrcid}}}

\newlist{questions}{enumerate}{2}
\setlist[questions,1]{label=\textbf{RQ}\arabic*.,ref=RQ\arabic*}
\setlist[questions,2]{label=(\alph*),ref=\thequestionsi(\alph*)}

\begin{document}
\title{When to intervene? Prescriptive Process Monitoring Under Uncertainty and Resource Constraints\thanks{Supported by the European Research Council (PIX Project).}}
%
\titlerunning{When to intervene? A PrPM method.}
%
\author{Mahmoud Shoush\orcidlink{0000-0002-7423-9909} \and
Marlon Dumas\orcidlink{0000-0002-9247-7476}}
\authorrunning{M. Shoush and M. Dumas}
%
\institute{University of Tartu, Estonia \\
\email{\{mahmoud.shoush, marlon.dumas\}@ut.ee}}

%
\maketitle              
%

\begin{abstract} \label{sec:abstract}
Prescriptive process monitoring approaches leverage historical data to prescribe runtime interventions that will likely prevent negative case outcomes or improve a process’s performance. A centerpiece of a prescriptive process monitoring method is its intervention policy: a decision function determining if and when to trigger an intervention on an ongoing case. Previous proposals in this field rely on intervention policies that consider only the current state of a given case. These approaches do not consider the tradeoff between triggering an intervention in the current state, given the level of uncertainty of the underlying predictive models, versus delaying the intervention to a later state. Moreover, they assume that a resource is always available to perform an intervention (infinite capacity). This paper addresses these gaps by introducing a prescriptive process monitoring method that filters and ranks ongoing cases based on prediction scores, prediction uncertainty, and causal effect of the intervention, and triggers interventions to maximize a gain function, considering the available resources. The proposal is evaluated using a real-life event log. The results show that the proposed method outperforms existing baselines regarding total gain.

\end{abstract}

\section{Introduction} \label{sec:introduction}
\textit{Prescriptive Process monitoring (PrPM)} is a family of process mining methods that trigger runtime actions to optimize a process's performance~\cite{fahrenkrog2021fire,metzger2020triggering}. PrPM methods use \textit{event logs} describing past business process executions to train \textit{machine learning (ML)} algorithms for two goals. \textit{First,} the trained ML models predict how an instance of the process (a.k.a.\ case) will unfold. For example,  whether the case leads to a positive outcome (e.g., a customer is satisfied) or a negative outcome (e.g., a customer launches a complaint)~\cite{teinemaa2019outcome}. 
\textit{Second}, PrPM methods use ML to assess  the effect of triggering an action (herein called an \textit{intervention}) on the probability of a negative outcome or a performance measure.

Recently, various PrPM methods have been proposed~\cite{shoush2021prescriptive,fahrenkrog2021fire,metzger2020triggering,bozorgi2021prescriptive}. These methods, however, implement intervention policies based on predictions of negative outcomes without considering  the uncertainty of these predictions. Also, they  trigger an intervention when the predicted probability of a negative outcome is above a threshold, without considering potential increases or decreases in this predicted probability that may occur as the case unfolds further. Finally, these methods do not consider the fact that there are limited resources available to perform the recommended interventions.




In this paper, we address the following problem: Given a set of cases, and given a type of intervention that generally decreases the likelihood of a negative case outcome. \textit{How do we select the cases for which applying the intervention (now or later) maximizes a gain function, considering the available resources to perform interventions?} Here, the gain function considers the tradeoff between the cost of applying the intervention to a case and the cost of negative outcomes.


To address this problem, we first apply an ensemble-based predictive model to estimate the negative outcome probability for each case, and we estimate the associated uncertainty. Using a causal model, we then determine the causal effect of applying an intervention on the negative outcome probability. We then use the negative outcome probability, the uncertainty, and the estimated causal effect and apply a \textit{filtering} and \textit{ranking} mechanism to identify cases for which an intervention would be most profitable (highest gain). We also consider the tradeoff (i.e., opportunity cost) between triggering an intervention in the current state, given the level of uncertainty of the predictive model, versus postponing the intervention to a later state. The paper reports an empirical evaluation comparing the proposed approach against state-of-the-art baselines.

The following section motivates the proposed method. Sect.~\ref{sec:bgrw} then presents background concepts and related work. Sect.~\ref{sec:approach} explains the proposed method, while Sect.~\ref{sec:evaluation} discusses the empirical evaluation. Finally, Sect.~\ref{sec:conclusion} concludes and discusses future work directions.



\section{Motivating Example} \label{sec:mex}
In \textit{a loan origination process}, a case starts when a customer submits his documents to obtain a loan. Then a process worker (or an employee) verifies the submitted documents. When they are valid, the employee sends an offer to the customer via different channels, such as phone calls or emails. This case ends positively when the customer accepts the offer and receives the loan, or negatively when the customer declines the offer or the employee rejects the application.

The principal concern arises when cases end negatively, leading to less payoff. One way to deal with this could be to predict negative cases based on prediction scores.  Then trigger an alarm to take a proactive action or an intervention, e.g., \textit{making a follow-up call}, when the prediction score exceeds a certain threshold. 

However, this strategy could be ineffective. Suppose an intervention policy where interventions are triggered to cases that are likely to end negatively based on low-quality prediction scores. Also, all employees are occupied and cannot immediately perform the intervention for all cases. Additionally, triggering interventions without considering their effect could be misleading since they may provide low or negative impact when utilized.

A more proper method is to quantify the prediction uncertainty to estimate how sure predictive models are with the prediction scores. Moreover, measuring the causal effect of utilizing interventions and considering the availability of resources. Another step that may enhance the overall payoff could be considering the tradeoff between triggering interventions now versus postponing them for a later state. In this paper, we discuss this method and evaluate its performance.

\section{Background and Related Work} \label{sec:bgrw}



\subsection{Predictive process monitoring}
Predictive process monitoring (PPM)~\cite{maggi2014predictive} is a complementary set of process mining methods to predict how ongoing cases will end. A PPM technique may, for instance, predict the remaining time for an ongoing case to be executed entirely~\cite{verenich2019survey}, the following action or activity to be executed~\cite{pauwels2021incremental}, or the outcome w.r.t group of outcomes, e.g., positive or negative~\cite{teinemaa2019outcome}. This paper focuses on the latter technique, known as an \textit{outcome-oriented} PPM.



Recent outcome-oriented PPM methods estimate the prediction scores, i.e., probability of negative outcomes, for ongoing cases and classify them positively or negatively. If the prediction scores exceed a threshold, e.g., above $0.5$, the ongoing case is considered more likely to end negatively. 

However, outcome-oriented PPM methods focus only on making predictions as accurate as possible, regardless of the quality of the predictions. These methods rely on several \textit{case bucketing techniques}~\cite{di2016clustering}, e.g., a single bucket where cases are made in the same bucket and train one ML algorithm instead of several. Also, they rely on various \textit{feature encoding} techniques~\cite{teinemaa2019outcome} to map each case into a feature vector to train the ML algorithm. For instance, an aggregate encoding in which all events from the beginning of the case are considered. Thus, several aggregate functions may be used to the values an event has carried throughout the case.  Also, a handful of possible \textit{inter-case} features are extracted~\cite{kim2022encoding} to enrich the training of ML or deep learning (DL) algorithms~\cite{kratsch2021machine}. Still, these techniques aim to improve the performance of the prediction scores and ignore quantifying the prediction quality via measuring \textit{prediction uncertainty}.

To the best of our knowledge, only one work from the literature considers estimating the model's prediction uncertainty explicitly, tackling another PPM task, i.e.,  remaining time~\cite{weytjens2021learning}. They learn the prediction uncertainty with artificial neural networks and a Monte Carlo (MC)~\cite{gal2016dropout} dropout technique that is unreliable for \textit{out-of-distribution} data (i.e., where there is an input case from a region very far from the trained data) and computationally expensive~\cite{abdar2021review}. 

Metzger et al.~\cite{metzger2020triggering} introduces an approach to measure  the reliability of prediction scores using an ensemble of DL classifiers at different process states. This approach does not discuss the estimation of the prediction uncertainty ignoring the situation of out-of-distribution input. Moreover, where predictive models provide several prediction scores for the same input, i.e., \textit{outcomes overlap}.




\subsection{Prescriptive process monitoring}
Prescriptive process monitoring (PrPM) methods go beyond predictions to prescribe runtime interventions to prevent or mitigate negative outcome effects. These methods aim to improve the performance of business process executions by determining if and when to trigger an intervention to maximize a payoff.

Diverse PrPM methods have been proposed. Metzger et al.~\cite{metzger2020triggering} suggest using ongoing cases prediction scores and their reliability estimate with a reinforcement learning technique to discover when to trigger runtime interventions. Another work by Fahrenkrog et al.~\cite{fahrenkrog2021fire} proposes triggering one or more alarms when cases are more likely to end negatively, followed by an intervention. 

Both the work of Metzger et al. and that of Fahrenkrog et al. identify cases that need intervention. They assume that resources are unbounded and consider only the current state of a given case to determine when to intervene. Instead, we study the tradeoff between intervening now or later based on the current and future prediction scores. Thus, we identify the most profitable case and assign resources to it, considering that resources are limited.

Weinzerl et al.~\cite{weinzierl2020prescriptive} suggest a PrPM method to recommend the next best activity from a list of possible activities with a higher preference for a pre-defined KPI. Khan et al.~\cite{kratsch2021machine} introduce a memory-augmented neural net approach to recommend the most suitable path (meaning a set of activities until the completion of the process) based on pre-specified KPIs. Both the work of Weinzerl et al. and that of Khan et al. do not discuss an exact idea of interventions or when to trigger them to maximize payoff.



\subsection{Causal Inference}


Causal Inference (CI)~\cite{xu2020causality} is a set of methods to predict what would occur if we adjust the process during its execution time by finding a cause-effect relationship between two variables, i.e., an intervention ($T$) and an outcome (Y). 

CI methods mainly unfold into two categories~\cite{guo2020survey}. The first category is \textit{structural causal models (SCMs)}, a multivariate statistical analysis method exploring structural relationships between dependent and independent variables. It depends mainly on discovering and building a causal graph by domain experts. 


The second category is \textit{potential outcome frameworks} (a.k.a., the Neyman-Rubin Causal Model). A statistical analysis method that does not require a pre-built causal graph like the SCMs and relies on the concept of potential outcomes. We use this category in this paper to automatically estimate the causal effect (or \textit{conditional average treatment effect (CATE)}) of intervention on negative cases instead of manually building causal graphs.

For example, in a loan-origination process, a customer would have a loan if he received an \textit{intervention (T)}, e.g., a follow-up call three days after receiving the first offer; otherwise, he would have a different \textit{outcome (Y)}, e.g., offer declined. Accordingly, to measure the $CATE$ of having a follow-up call, we need to compare $Y$ for the same customer when receiving the follow-up call, i.e., $T=1$, and not receiving the follow-up call, i.e., $T=0$.

Recent work uses the potential outcome method to estimate the $CATE$ of utilizing interventions. Specifically, in~\cite{bozorgi2021prescriptive}, the Authors introduce a PrPM technique to measure the effect of intervention at an individual case level to reduce the cycle time of the process.  It targets another PPM problem and considers only one process execution state; it also assumes that resources are unbounded.

In our previous work~\cite{shoush2021prescriptive}, we propose a PrPM method that utilizes the potential outcome method and a resource allocator technique in the outcome-oriented PPM to allocate resources to cases with max gain. That work considers only a given case’s current state and triggers interventions when the prediction score and the causal effect exceed a threshold. However, we did not discuss the tradeoff between triggering an intervention now, given the level of uncertainty of the underlying predictive models versus later.

\section{Approach} \label{sec:approach}

\begin{figure*}[!htb]
	\begin{center}
        \resizebox{\textwidth}{!}{\includegraphics{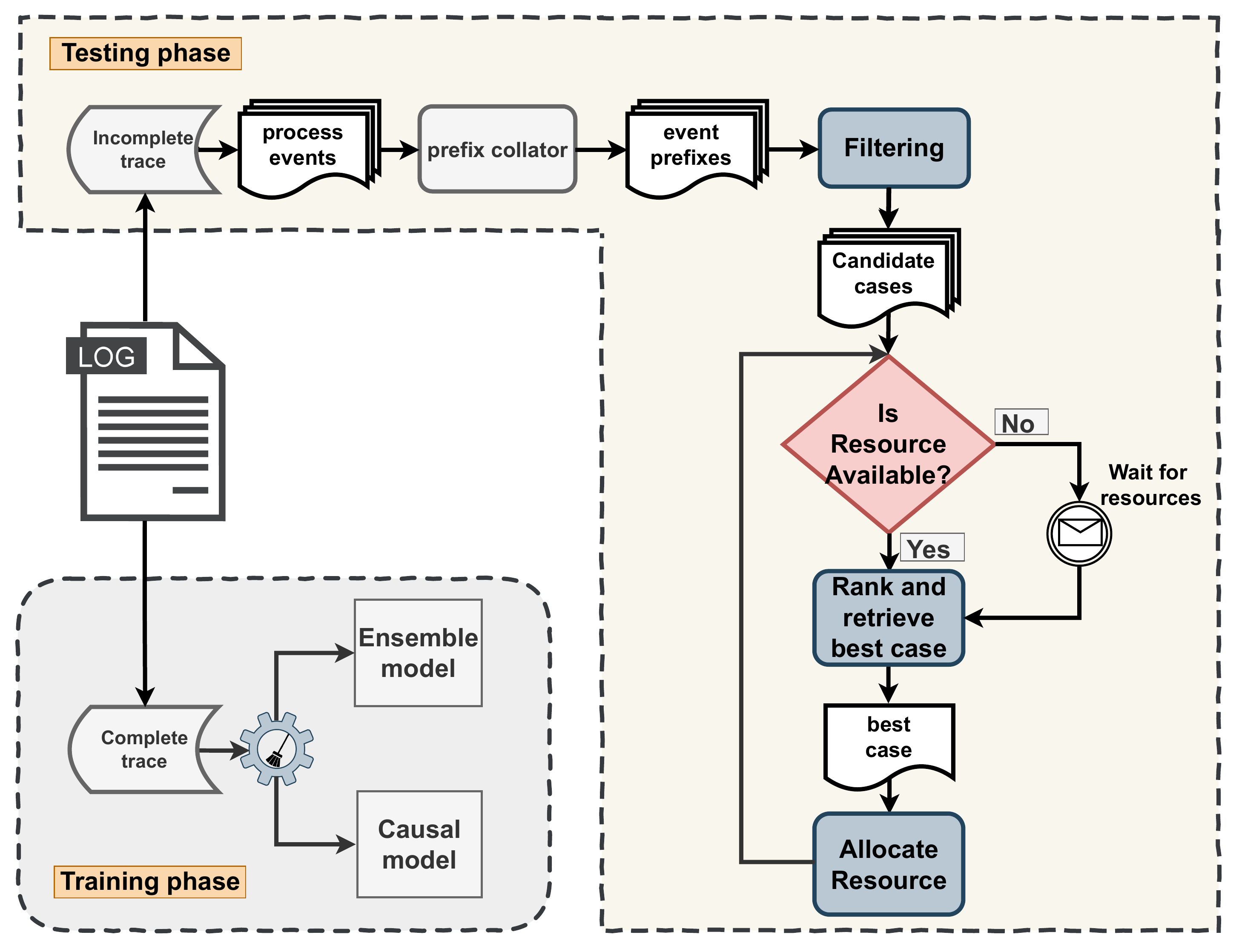}}
		\caption{An overview of the proposed approach.}
		\label{fig:approach}
	\end{center}
\end{figure*}

The proposed PrPM method consists of two main phases, training and testing, see Fig.~\ref{fig:approach}. In the training phase, we train two ML models, i.e., predictive and causal. While in the testing phase, we present filtering and ranking techniques to determine the most profitable case. Then, decide when to trigger an intervention for the selected case to maximize a gain function—considering ongoing cases’ current and future state scores, uncertainty estimation, and resource availability.

\subsection{Training Phase}
We first prepare the process execution data; then, we construct an ensemble-based predictive model to estimate the prediction scores, i.e., the probability of cases likely to end negatively and quantify the prediction uncertainty. Further, we build a causal model to estimate the $CATE$.





\subsubsection{Event Log Preprocessing}
\begin{figure*}[!htb]
	\begin{center}
        \resizebox{\textwidth}{!}{\includegraphics{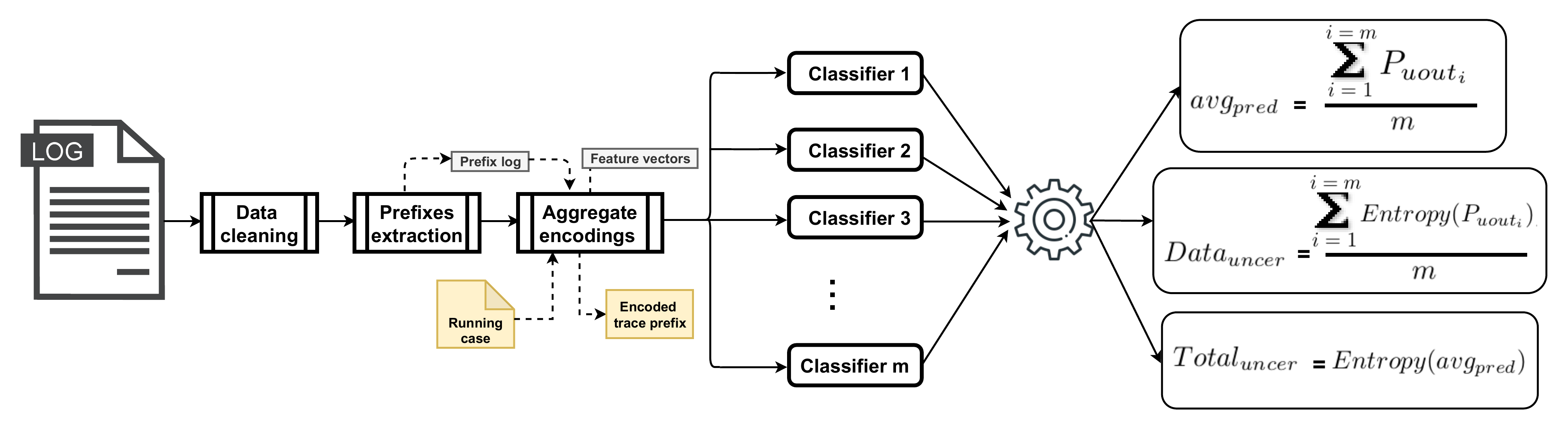}}
		\caption{An overview of the ensemble model. }
		\label{fig:pred}
	\end{center}
\end{figure*}

This step is vital for PPM or PrPM tasks. In PPM, it includes \textit{data cleaning, prefix extraction, and prefix encoding} see Fig.~\ref{fig:pred}. These steps have been discussed in Teinemaa et al.~\cite{teinemaa2019outcome}, and we follow their suggestions here. We first pre-process the log to dismiss incomplete cases and then extract length $k$ from every case that results in a so-called \textit{prefix log}. This prefix extraction ensures that our training data is equivalent to the testing data. Finally, we encode each trace prefix into a feature vector ($X$) to train the predictive ensemble model, see Fig.~\ref{fig:pred}.



While in PrPM, one further step is needed to analyze and understand the data and the business objective to identify an intervention $T$ that could positively impact an outcome $Y$. Moreover, determining what other variables ($W$: a.k.a. \textit{confounders}) affect the intervention and outcome. 





\subsubsection{Ensemble model}
We construct an outcome-oriented predictive model (a classification problem from an ML perspective) via ensemble learning~\cite{dietterich2002ensemble}, as shown in Fig. \ref{fig:pred}. The principal assumption of ensemble learning is constructing one robust predictive model from several weak ones. Accordingly, the overall prediction scores performance would be superior where overfitting and the chance of getting a local minimum are avoided, which has two advantages to our work. First, it ensures  that we will accurately predict the probability of negative outcomes, i.e., \textit{the prediction score}. Second, it allows estimating \textit{the prediction uncertainty}.




A single classifier in the ensemble is a  probabilistic model $cls_i: f(X, P_{uout_i})$. $X$ is the input feature vector, and the $P_{uout_i}$ is the estimated probability of cases likely to end with negative outcomes where $i \in \{1, m\}$, and $m$ is the number of classifiers in the ensemble. We then define the \textit{prediction score} as the average of individual classifiers’ prediction scores $(avg_{pred})$, as shown in Eq.~\ref{eq:avgpred}. The following step is to estimate the prediction uncertainty.





	\begin{equation}
	avg_{pred} = \frac{\sum_{i=1}^{m}P_{uout_i}}{m}
        \label{eq:avgpred}
    \end{equation}
    
There are two sources of uncertainties where predictive models become unsure about predictions~\cite{hullermeier2021aleatoric}. The first source is \textit{data (or aleatoric)} uncertainty $(\sigma)$. It is a property of the distribution that generates cases, and it occurs when \textit{outcomes overlap} or there is \textit{noise} in the underlying data distribution. The second source is \textit{knowledge (or epistemic)} uncertainty $(\rho)$. It is a property of the predictive model's learning parameters and arises due to a lack of model knowledge. It appears when there is \textit{out-of-distribution} input. Both sources form the prediction uncertainty, i.e., \textit{total uncertainty} ($total_{uncer}$), see Eq.~\ref{eq:tuncer1}.

    

	\begin{equation}
	total_{uncer} = \sigma + \rho
	\label{eq:tuncer1}
    \end{equation}

The proper level of $\sigma$ is defined as \textit{entropy} of the actual underlying data distribution~\cite{malinin2020uncertainty}. Here, the entropy is the average level of surprise or uncertainty inherent in the possible outcomes. It is calculated for a random variable $S$ with $c$ in $C$ discrete states, see Eq.~\ref{eq:entropy}, where $P(c)$ is the probability of $c$ to occur. 

	\begin{equation}
	Entropy(S) = - {\sum_{c \in C} P(c)\log(P(c)}
	\label{eq:entropy}
    \end{equation}

However, we do not have access to the actual underlying data distribution, but our model is probabilistic and is trained on this data. So measuring the entropy of our probabilistic model, i.e., trained using \textit{negative log-likelihood}, estimates the level of $\sigma$. In particular, we obtain a distribution over outcome labels from individual classifiers in the ensemble for a given case, and $\sigma$ is the average \textit{entropy} of the individual prediction scores, as shown in Eq.~\ref{eq:duncer}.
    


	\begin{equation}
	\sigma = \frac{\sum_{i=1}^{m}Entropy(P_{uout_i})}{m}
        \label{eq:duncer}
    \end{equation}

Furthermore, we rely on the \textit{Bayes rule}~\cite{stone2013bayes} to estimate the $\rho$~\cite{malinin2020uncertainty}. Assume we obtain a posterior over model parameters that give us the distribution over likely models that have generated the data. Accordingly, models sampled from the obtained distribution agree on data they have seen and provide similar predictions, indicating low $\rho$. However, if the models do not understand the input, they provide diverse predictions and strongly disagree, indicating high $\rho$. Thus, $\rho$ is the mutual information between the models' parameters and the predictions. 




Similarly, the total uncertainty is the entropy of the average prediction, see Eq.~\ref{eq:tuncer2}. To exemplify the estimation of the prediction uncertainty ($total_{uncer}$), suppose we show the ensemble of classifiers several kinds of input. (1) We give input that all classifiers understand and yield the exact prediction scores. Consequently, classifiers are confident about their prediction scores, and the $total_{uncer}$ is minimum. (2) We show input that all classifiers understand and generate identical predictions but high entropy distribution over outcomes; then, classifiers are uncertain with high $\sigma$. (3) We show the ensemble something none of the classifiers understand; hence, all classifiers yield different prediction scores because the input comes from a very far region from the training data. Thus, the ensemble is very diverse with high entropy because we average various probability distributions together, then classifiers are uncertain with high $\rho$.

	\begin{equation}
	total_{uncer} = Entropy (avg_{pred})
        \label{eq:tuncer2}
    \end{equation}

In addition to the ensemble model that estimates $avg_{pred}$ and $total_{uncer}$ in the training phase, we train a causal model to measure the $CATE$. In particular, we utilize an \textit{orthogonal random forest (ORF)} algorithm because it reasonably deals with high-dimensional variable spaces. Thus, it is beneficial since the process execution logs have numerous event attributes with categorical values.

Estimating the $CATE$ means we evaluate the difference between the probability of a negative outcome if we intervene and if we do not intervene. The higher the differences, the stronger the effect of the intervention. For explanation, we recall the motivating example in Sect.~\ref{sec:mex}, where the goal was to improve the performance of a loan origination process by raising the number of successful applications. Accordingly, we would give customers who are likely to decline the first offer another offer in a way that affects the probability of declining the first offer positively. Then we estimate what would happen for the probability of negative outcome when we send customers a second offer and when we do not. 





\subsection{Testing Phase}

We first use the trained ensemble and causal models to obtain $avg_{pred}$, $total_{uncer}$, and $CATE$ scores for ongoing cases. Then, we use these scores to \textit{filter} ongoing cases into candidate ones and \textit{rank} them to choose the most profitable case to maximize \textit{a gain} function. We consider that resources are bounded and compare ongoing cases' current and future state scores. 


\subsubsection{Filtering}

At run time, new events of ongoing cases keep coming continuously, and we first collate them via \textit{a prefix collator} to accumulate the sequence of events. Thus, at any point in time, we could have one or multiple ongoing cases that we choose from which one is the most profitable to trigger the intervention. Hence, ongoing cases need to be filtered to minimize the search space.


We use $avg_{pred}$, $total_{uncer}$, and $CATE$ scores to filter ongoing cases into candidate ones. The essence of these scores varies from one to another. The $avg_{pred}$ gives information about whether ongoing cases are likely to end negatively or not. At the same time, the $total_{uncer}$ shows how sure the model is with its predictions. It ranges from $0$ to $1$, where the predictive model is entirely certain or uncertain, respectively. Moreover, the $CATE$ is crucial to any PrPM technique, representing the expected impact of utilizing intervention on an ongoing case, e.g., when $CATE$ is above $0$, it impacts positively.

All the abovementioned scores ($avg_{proba}$, $total_{uncer}$, $CATE$) are vital to determining candidate cases from which we will choose the most profitable. However, there are two other critical aspects to determine when to intervene and to define the most profitable case; how the estimated scores will change in the following state and whether resources are available or not.


\subsubsection{Future state scores estimation}
Considering only the current state scores of ongoing cases regardless of investigating what would occur in the future could be misleading. Because if we decide not to intervene, maybe it will be more effective to achieve higher gain when we utilize the intervention later than now, or we will be more sure about the prediction of the outcome and decide not to intervene at all. So, discovering what will happen in the future of ongoing cases allows deciding whether to intervene now or later. 


To estimate what will happen in the future state, we predict the $avg_{pred}$, $CATE$, and $total_{uncer}$ in the future. So for each score, we will have two values: one representing the current state ($c\_avg_{pred}$, $c\_CATE$, $c\_total_{uncer}$) and the other representing the future ($f\_avg_{pred}$,  $f\_CATE$, $f\_total_{uncer}$). We follow a technique inspired by the \textit{k-nearest neighbors (KNN)}~\cite{peterson2009k} algorithm to get scores representing the future state, given the degree of \textit{similarity} and \textit{frequency}. 



We look at previous cases similar to the ongoing one at the next prefix. For example, in a given case at prefix $4$, we want to know what will happen at prefix $5$. Then, we capture similar prefixes and define an aggregate score, i.e., the \textit{weighted average}. We consider the degree of \textit{similarity} using Euclidean distance to the current prefix, their \textit{frequency} because higher frequency means more weight and the scores for all similar prefixes. Accordingly, scores representing the future state ($f\_avg_{pred}$,  $f\_CATE$, $f\_total_{uncer}$) of ongoing cases are the weighted average from similar previous prefixes. The next step is to use current and future estimates to rank candidate cases and select the most profitable. 


\subsubsection{Rank and retrieve the best case.}

To define the most profitable case, we first distinguish between \textit{gain} and \textit{adjusted gain}. The \textit{gain} is the benefits we attain at one state only, either current or future, and it means we estimate two gains, one for the current state ($c\_gain$) and the other for the future state ($f\_gain$). In contrast, the \textit{adjusted gain} ($adj_{gain}$) is the benefits we attain considering current and future states. We use the $adj_{gain}$ as a decision function to determine whether to intervene now or later.


Triggering interventions may come with benefits and, at the same time, comes at a cost. The costs can vary from one process to another. However, for a given case $c_{id}$, there is generally a cost for applying the intervention ($cost(c_{id}, T_{i=1})$)) when $T=1$ and not applying the intervention ($cost(c_{id}, T_{i=0})$) when $T=0$. 



The cost of not applying the intervention describes how much we lose if the negative outcome occurs, and it relies on the $avg_{pred}$ and cost of negative outcomes ($c_{uout}$), as shown in Eq.~\ref{eq:cnt}. In contrast, the cost of utilizing the intervention refers to how much we decrease the probability of negative outcomes considering the intervention cost ($c_{T_1}$), see Eq.~\ref{eq:ct}, which assumes $CATE$ is reliable. Costs are mainly identified via domain knowledge; however, we assume that the $c_{T_1}$ is less than the $c_{uout}$ to obtain meaningful results. 


	\begin{equation}
        cost(c_{id}, T_{i=0}) = avg_{pred} * c_{uout}
        \label{eq:cnt}
    \end{equation}
	\begin{equation}
        cost(c_{id}, T_{i=1}) = (avg_{pred} - CATE_1) * c_{uout} + c_{T_1}
        \label{eq:ct}
    \end{equation}
    
The corresponding gain ($gain(c_{id}, T_{i=1})$) from utilizing the intervention on $c_{id}$ is the benefits that allow the highest cost reduction in Eq.~\ref{eq:gain}




	\begin{equation}
        gain(c_{id}, T_{i=1}) = cost(c_{id}, T_0)  - cost(c_{id}, T_{i=1})
        \label{eq:gain}
    \end{equation}



    

We estimate the current state gain ($c\_gain(c_{id}, T_{i=1})$) using scores from the ensemble and causal models, see Eq.~\ref{eq:cgain}. In contrast, the gain for the future state ($f\_gain(c_{id}, T_{i=1})$) is based on the weighted average scores from previous similar prefixes, see Eq.~\ref{eq:fgain}.


	\begin{equation}
        c\_gain(c_{id}, T_{i=1}) = c\_cost(c_{id}, T_0)  - c\_cost(c_{id}, T_{i=1})
        \label{eq:cgain}
    \end{equation}
    	\begin{equation}
        f\_gain(c_{id}, T_{i=1}) = f\_cost(c_{id}, T_0)  - f\_cost(c_{id}, T_{i=1})
        \label{eq:fgain}
    \end{equation}
    
Determining the gain for candidate cases' current and future states is vital to define the adjusted gain. To explain the adjusted gain, we first define \textit{an opportunity cost} that measures what we lose when choosing between two or more alternatives—for example, utilizing the intervention now or later. The opportunity cost ($opp_{cost}$) is the difference between the gain we could achieve in the future state of a given case and the gain in the current state, as shown in Eq.~\ref{eq:oppc}.


	\begin{equation}
        opp_{cost} = f\_gain(c_{id}, T_{i=1})  - c\_gain(c_{id}, T_{i=1})
        \label{eq:oppc}
    \end{equation}

Given the opportunity cost, we define the \textit{adjusted gain} as the payoff (or gain) we acquire from utilizing the intervention on candidate cases considering current and future states. It is the difference between the $c\_gain(c_{id}, T_{i=1})$ and the $opp_{cost}$, as shown in Eq.~\ref{eq:again}. Thus, we define \textit{the most profitable case} as the one with the highest $adj_{gain}$, which means the lowest $opp_{cost}$.


	\begin{equation}
        adj_{gain} = c\_gain(c_{id}, T_{i=1})  - opp_{cost}
        \label{eq:again}
    \end{equation}

\begin{table}[hbpt]
\centering

\caption{An example of defining gain.}
	\label{tab:againex}
\begin{tabular}{cccccl}
\hline
$c_{id}$ \hspace{0.3cm} & $c\_gain(c_{id}, T_{i=1})$ \hspace{0.3cm} & $f\_gain(c_{id}, T_{i=1})$ \hspace{0.3cm} & $opp_{cost}$ \hspace{0.3cm} & adju \hspace{0.3cm} & \multicolumn{1}{c}{decision} \\ \hline
A   & 7     & 12     & 5   & 3    & Wait                         \\
B   & 5    & 1    & -4  & 9   & Treat                        \\
C   & 3     & 3     & 0   & 3    & Neutral     \\ \hline                
\end{tabular}
\end{table}

For example, suppose we filtered ongoing cases into three candidates (see Tab.~\ref{tab:againex}) eligible for the intervention. Also, there is an available resource to do the intervention; we need to choose which one is more suitable to intervene now or later. If we consider only the gain from the current state, we assign resources to $c_{id}=A$ and treat it now. However, If we think about the gain from the future state, we observe that later we can achieve more gain if we do not intervene and previously assigned resources to $c_{id}=A$ inaccurately. Hence, it is more beneficial to allocate resources to $c_{id}=B$ and apply the intervention now since we might lose current gain later. Thus, using the adjusted gain to decide when to intervene could enhance the performance of PrPM.

Selecting the best or most profitable case can be judged efficiently based on the adjusted gain to maximize the total gain. However, triggering interventions as often as we want and immediately is impossible since resources are bounded in practice, which is the last aspect we need to consider.

\subsubsection{Resource Allocator.}
Monitoring resources and assigning them to cases that need intervention is critical. The resource allocator checks the availability of resources. Once the most profitable case is selected and a free resource is available, we assign that resource to the selected case and block it for a certain time, i.e., \textit{treatment duration} ($T_{dur}$).  The number of available resources and the time required to perform the intervention could be identified via domain knowledge~\cite{shoush2021prescriptive}. 



\section{Evaluation} \label{sec:evaluation}
To verify the effectiveness and relevance of the proposed method, we experimentally investigate whether we can learn when to trigger an intervention to maximize the total gain at the run time, considering the tradeoff between intervening now or later. We compare our results to baselines that consider either predictive models without quantifying the prediction uncertainty~\cite{fahrenkrog2021fire,metzger2020triggering} or only the current case's execution state scores~\cite{shoush2021prescriptive,bozorgi2021prescriptive} as state-of-the-art baselines by addressing the following research questions:  


    \begin{questions}
        \item To what extent does taking into account the current uncertainty prediction enhance the total gain?\label{rq:rq1}
        
        \item To what extent does taking into account the derivative of the uncertainty prediction and the adjusted gain enhance the total gain?\label{rq:rq2}
    \end{questions}
    
\subsection{Dataset}

We use a real-life event log, named \emph{BPIC2017}\footnote{\url{https://doi.org/10.4121/uuid:5f3067df-f10b-45da-b98b-86ae4c7a310b}}, publicly available from the 4TU.ResearchData. The log describes the execution of a loan origination process, and we choose this log for several reasons. First, it contains a clear notion for outcome definition and interventions utilized to show our method's efficiency. Second, this log contains $31,413$ applications and $1,202,267$ events, which is large enough and frequently used for predictive and prescriptive methods. 

The \textit{BPIC2017} log is characterized by various case and event attributes, and we include all of them in our experiments. Additionally, we extracted other essential critical features in our work, such as the number of sent offers, event number, and additional temporal features.

We used all original and extracted attributes as input for ensemble and causal models in the preprocessing step. Then we defined case outcomes based on the end state of cases, i.e., \textit{“A\_Pending”} state means positive outcome and \textit{“A\_canceled} or \textit{A\_Denied”} means negative outcome. After that, we defined the intervention that positively impacts the negative outcome as sending a second offer (or \textit{"Creat\_Offer"} activity) to all customers who received only one offer. Accordingly, we denoted cases with $T = 1$ based on the number of offers sent to each, i.e., cases that receive only one offer. Then, we extracted length prefixes no more than the $90^{th}$ percentile for each case to avoid bias from lengthy cases. Finally, we used an \textit{aggregate encoding} to encode the extracted prefixes.


\subsection{Experimental Setup}
The experiments show our method's effectiveness during operation time with two main objectives: deciding when to intervene, either now or later, and selecting the most profitable case among all candidates to maximize the \textit{total gain}.



We adapted an ensemble model based on a \textit{Gradient Boosting Decision Tree (GBDT)} method to estimate the $avg_{pred}$ and $total_{uncer}$. In particular, we used  \textit{Catboost}~\cite{prokhorenkova2018catboost}, an open-source GBDT library with several tools to quantify the prediction uncertainty and automatically handle categorical features.


\textit{Catboost} is trained with a \textit{negative log-likelihood} loss and \textit{Langevin optimization}~\cite{ustimenko2021sglb} to ensure global conversion instead of a local optimum and generate an ensemble of several independent GBDT. Additionally, we used the following parameters during training: ensemble size of $50$, a learning rate of $0.05$, a subsample of $0.82$, and a max tree depth of $12$.

Catboost returns a probability distribution over the case outcomes. This
distribution is based on a given model version, i.e., on the seed used to initialize the model parameters before training. So we train the same model using different seed initialization and evaluating these models on the same input to obtain the $avg_{pred}$ and its $total_{uncer}$. We use an ORF algorithm implemented in the  \emph{EconMl}\footnote{\url{https://github.com/microsoft/EconML}} to train a causal model to estimate the CATE.





We follow the machine learning workflow to train both ensemble and causal models. We temporally split the data into training ($60\%$), validation ($20\%$ ), and testing ($20\%$) sets. Training and validation sets are used to train and tune model parameters, while the testing set is used to evaluate the model’s performance.
\begin{table}[hbpt]
\centering
\caption{Parameter settings of the introduced method}
\label{tab:param}
\resizebox{\textwidth}{!}{
\begin{tabular}{cccccccc}
\hline
$avg_{proba}$\hspace{0.5cm} & $CATE$\hspace{0.5cm}  & \hspace{0.5cm}$total_{uncer}$ & $\Delta total_{uncer} \hspace{0.5cm}$                                            & \hspace{0.5cm}\#resources \hspace{0.5cm} & \hspace{0.5cm} $c_{uout}$ \hspace{0.5cm}  \hspace{0.5cm}& $c_{T_1}$ \hspace{0.5cm}& $T_{dur}$  (sec)                                                                       \\ \hline
\textgreater $0.5$          & \textgreater $0$      & $< 0.25, 0.5, 0.75$ \hspace{1cm}     & $< 0, -0.5, -0.1, -0.15, -0.2, -0.25, -0.3$ \hspace{0.3cm} & $1, 2, ...10$ & $20$    & $1$   & \begin{tabular}[c]{@{}c@{}}Fixed = $60$\\  \hspace{0.8cm}Normal $\in$ \{$1, 60$\}\\ \hspace{1.4cm}  Exponential $\in$ \{$1, 60$\}\end{tabular}
\\ \hline
\end{tabular}}
\end{table}

During the testing time, we follow the configurations shown in Tab.~\ref{tab:param}. First, ongoing cases are filtered into candidates. We filter cases based on the estimated probability of $avg_{pred}$ \textgreater $0.5$ to ensure that cases are highly probable to end negatively and $CATE$ \textgreater $0$ to guarantee that intervention has a positive impact.


Additionally, we filter cases using the $total_{uncer}$ to see how sure the predictive model is with the predictions. The estimated $total_{uncer}$ ranges from $0$, where the model is certain, to $1$, which is entirely uncertain. We experiment with three thresholds (see Tab.~\ref{tab:param}) when considering only the current state of cases. 

On the other hand, we use the $opp_{coast}$ or $adj_{gain}$ when we consider both current and future states of ongoing cases. At the same time, with and without a derivative of the total uncertainty ($\Delta total_{uncer}$), representing the difference between $c\_total_{uncer}$ and $f\_total_{uncer}$. Accordingly, we filter cases based on the $c\_avg_{pred}$ and $c\_CATE$ and then select the case with the highest $adj_{gain}$ with and without $\Delta total_{uncer}$ when it is below $0$ or negative values (see Table~\ref{tab:param}), which means the predictive model becomes more uncertain in the future state.






\subsection{Results}
We show results here based on one $T_{dur}$ distribution, i.e., fixed. Since the normal distribution achieves similar results to the fixed in terms of the total gain and is higher than the exponential distribution as the variability is higher in the latter. Also, we observed that the behavior for different thresholds for $total_{uncer}$ and $\Delta total_{uncer}$ is the same, i.e. when we lower the threshold, fewer cases are treated with no substantial effect on the total gain per case. Hence, we present results using one threshold, i.e., the higher value for $total_{uncer} < 0.75$ and $\Delta total_{uncer} < 0$. However, The full results of experimenting with all thresholds and $T_{dur}$ distributions are available in supplementary material~\footnote{\url{https://zenodo.org/record/6381445\#.YjwaFfexWuA}{\label{supp}}}.

\begin{figure}[!h]
\centering
\begin{subfigure}{0.5\textwidth}
  \centering
  \includegraphics[width=\linewidth, ]{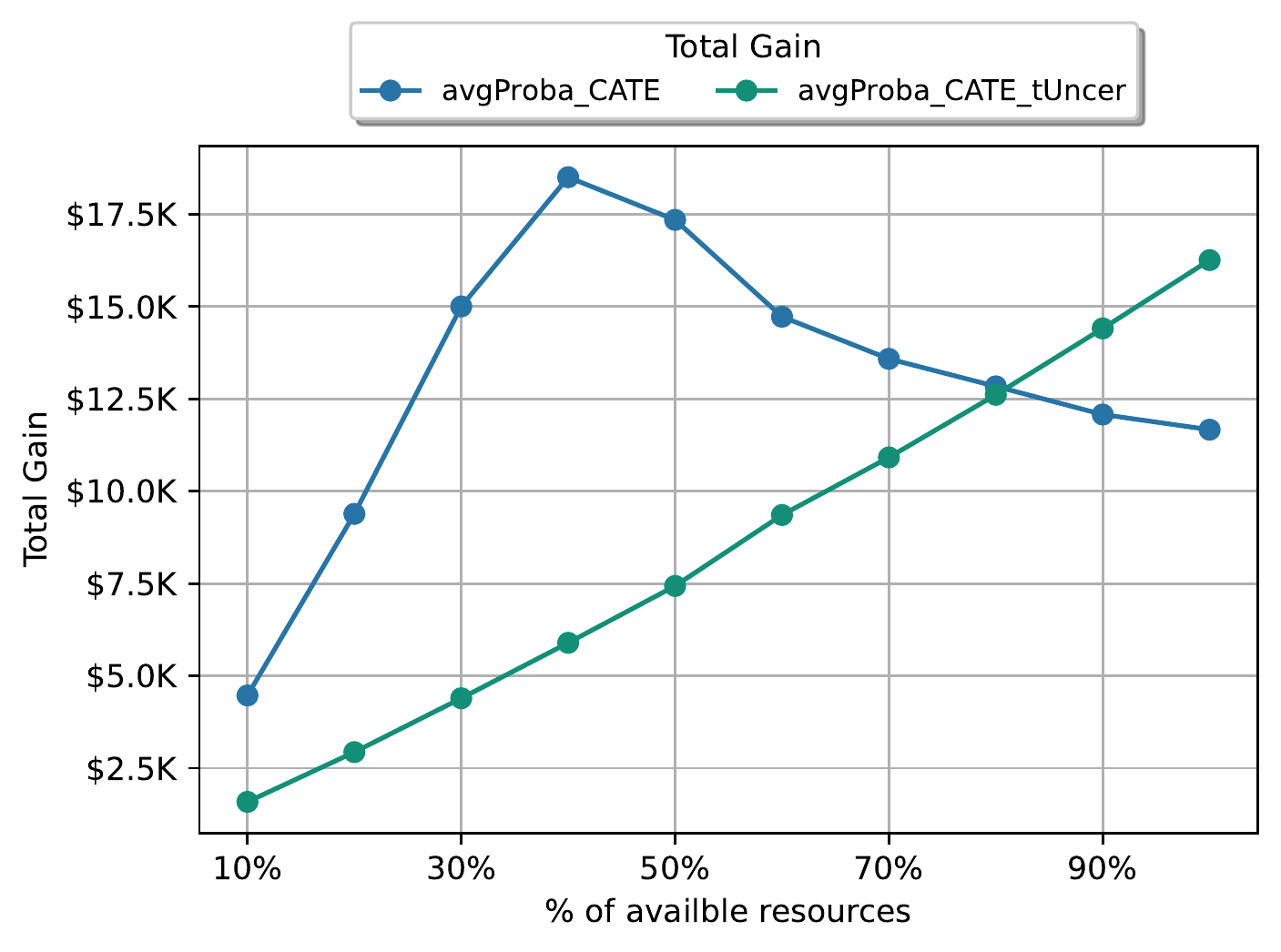}
  \scalebox{0.7}{\parbox{\linewidth}{%
  \caption{RQ1: Where $avgProba\_CATE$ means $c\_avg_{proba} > 0.5$ \& $c\_CATE > 0$. Also, $avgProba\_CATE\_tUncer$ means  $c\_avg_{proba} > 0.5$ \& $c\_CATE > 0$ \&  $c\_total_{uncer} < 0.75$.}
  \label{fig:rq1}}}
\end{subfigure}%
\begin{subfigure}{0.5\textwidth}
  \centering
  \includegraphics[width=\linewidth]{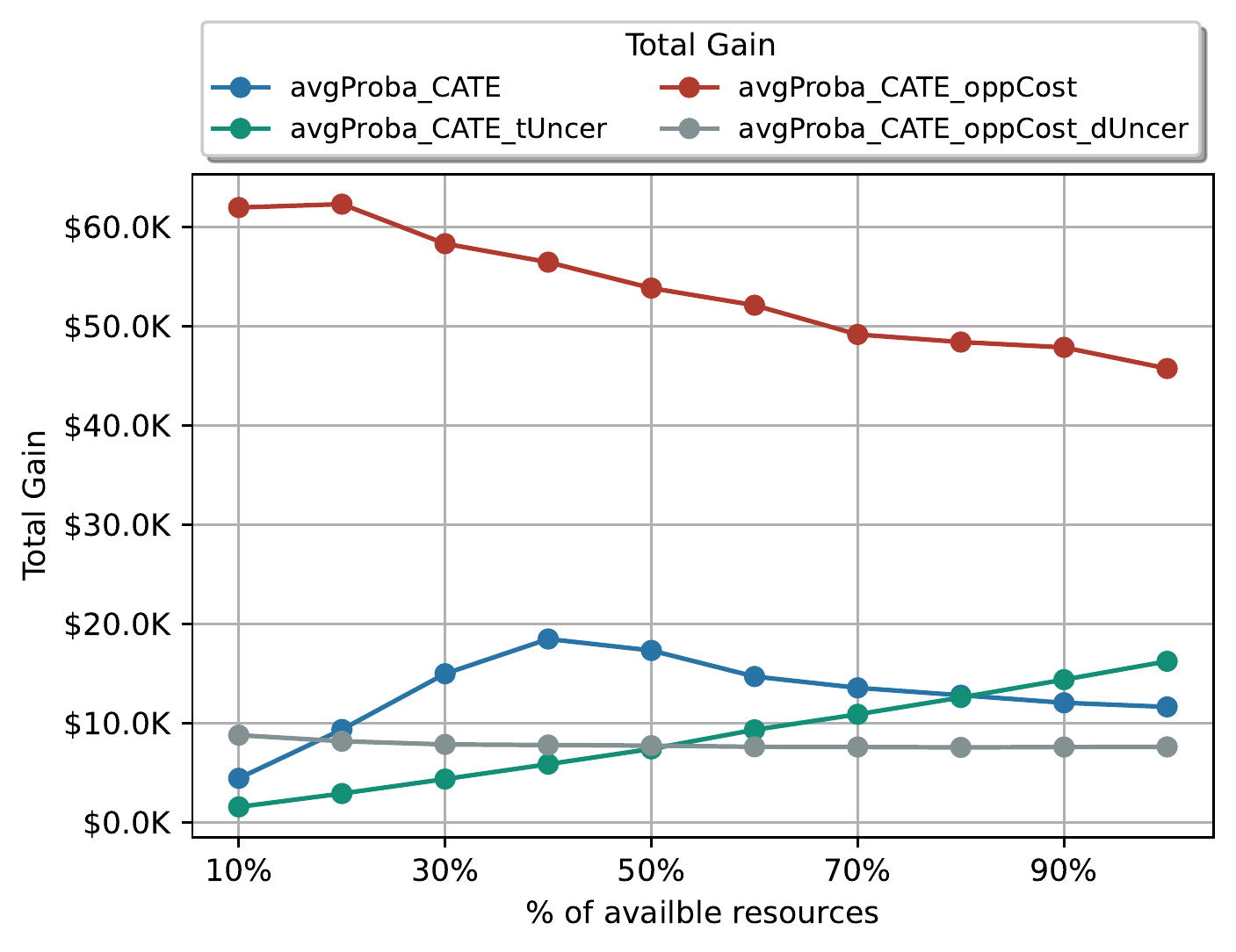}
  \scalebox{0.7}{\parbox{\linewidth}{%
  \caption{RQ2: Where $avgProba\_CATE\_oppCost$ refers to the $adj_{gain}$ is maximized. Also, $avgProba\_CATE\_oppCost\_dUncer$ refers to the $adj_{gain}$ is maximized \& $\Delta total_{uncer} <0$.}
  \label{fig:rq2}}}
\end{subfigure}
\caption{The total gain progress under the same number of available resources.}
\label{fig:res}
\end{figure}

To discuss~\ref{rq:rq1}, we consider only the current state scores of ongoing cases. Then, examine how the total gain evolves when adding the  current prediction uncertainty ($c\_total_{uncer}$) to the filtering step. Besides the estimated prediction score ($c\_avg_{proba}$ ) and the causal effect ($c\_CATE$). We compare this to baselines shown in~\cite{shoush2021prescriptive,bozorgi2020process}, where only the $avg_{proba}$ and $CATE$ are set, see Fig.~\ref{fig:rq1}.

The results in Fig.~\ref{fig:rq1} explain that adding the current prediction uncertainty improves the total gain when the available resources exceed 80\%. Also, the baseline treats more cases with less total gain (as shown in the supplementary~\ref{supp}). In contrast, adding the current prediction uncertainty allows efficient allocation of resources since fewer cases are treated but with a higher total gain than the baseline. Because considering the current prediction uncertainty does not trigger the intervention until the predictive model becomes more sure about its predictions and achieves higher total gain.


Turning to discuss~\ref{rq:rq2}, we consider two states’ scores of ongoing cases, i.e., current and future, to estimate the $opp_{cost}$ and the $adj_{gain}$ and their impacts on the total gain. Hence, we decide when to intervene and allocate the available resources to ongoing cases based on two things: first, cases with the maximum $adj_{gain}$ only, and second, cases with the maximum $adj_{gain}$ and the $\Delta total_{uncer} <0$. We compare this method to the baseline in~\ref{rq:rq1} and consider the current prediction uncertainty only, see Fig.~\ref{fig:rq2}.


The results in Fig.~\ref{fig:rq2} show that the total gain is highly affected by the trade off of current and future state scores under the same quantity of consumed resources. The proposed method achieves a significant improvement in terms of the total gain and outperforms the baselines (cf.~Fig \ref{fig:rq1}). We found that considering the $adj_{gain}$ as a decision function to determine when to intervene is more efficient than considering it with the $\Delta total_{uncer}$. Still, considering the derivative of the prediction uncertainty could achieve a higher gain when the available resources are below 10\%. Because when the number of available resources increases, more cases are treated without ensuring that the $total_{uncer}$ is sufficient.

The proposed PrPM method results in Fig.~\ref{fig:res} demonstrate a higher total gain than the baselines. Accordingly, estimating the prediction uncertainty of the underlying predictive model and analyzing the tradeoff between triggering an intervention now versus later based on the adjusted gain and the opportunity cost can optimize the performance of PrPM methods, hence, business processes.



\subsection{Threats to Validity}
Our method's evaluation has an external validity threat (lack of generalizability) because of its dependence on only one dataset. Accordingly, the evaluation is vague and needs more experiments using other logs to be followed up.

We assume that $CATE$ is accurate and will reduce the probability of negative outcomes. Also, the intervention will be triggered only once for each ongoing case. Accordingly, there is a threat to ecological reality where cases may be treated more than once via multiple interventions. Moreover, the  $CATE$ may not represent the natural causal effect because of unobserved confounders.

The experiments’ setup used one feature encoding technique and did not discuss the selection bias of the causal model due to variables that affect both the outcome and the intervention. Utilizing other encoding and selection bias techniques is a hint for future work to improve the performance of PrPM methods.
\section{Conclusion and Future Work} \label{sec:conclusion}

We presented a prescriptive monitoring method to determine if and when an intervention should be triggered on ongoing cases to maximize a gain function. The method leverages an ensemble model to estimate the probability of negative case outcomes and the associated prediction uncertainty. They are combined with a causal model to assess the effect of an intervention on the case outcome. These estimates and a filtering and ranking method are embedded in a resource allocator. It assigns resources to perform interventions to maximize total gain, considering the tradeoff between intervening now or later. An initial evaluation shows that taking into account the possible future states of ongoing cases and the level of prediction uncertainty leads to a higher gain than baselines that rely only on the current state of each case.

The proposed prescriptive method does not include constraints on when an intervention may be triggered. In practice, it is often not possible to trigger an intervention at any point in a case. For example, sending a second loan offer to a customer who has already accepted an offer at another bank is counter-productive. A direction for future work is to extend the approach with temporal rules on the interventions and  consider these rules in the intervention policy.

The proposed method is limited to handling one type of intervention. Also, an intervention is applied at most once to a case. Lifting these restrictions is another direction for future work.



\medskip\noindent\textbf{Reproducibility.} The implementation and source code of the method, together with instructions to reproduce the evaluation, can be found at:  \url{https://github.com/mshoush/prescriptive-monitoring-uncertainty}.

\bibliographystyle{splncs04}
\bibliography{Ref}







\end{document}